\documentclass{article}

\PassOptionsToPackage{numbers, compress}{natbib}

\usepackage[preprint]{neurips_2024}
\usepackage{amsmath}
\usepackage{float}

\usepackage[utf8]{inputenc} 
\usepackage[T1]{fontenc}    
\usepackage{hyperref}       
\usepackage{url}            
\usepackage{booktabs}       
\usepackage{amsfonts}       
\usepackage{nicefrac}       
\usepackage{microtype}      
\usepackage{xcolor}         

\usepackage{comment}
\usepackage{csquotes}
\usepackage{graphicx}
\usepackage{tikz}

\usepackage{ifthen}

\title{Human-like Forgetting Curves in Deep Neural Networks}

\author{%
  Dylan Kline  \\
  Hajim School of Engineering \& Applied Sciences: Department of Computer Science \\
  University of Rochester\\
  \texttt{dkline4@u.rochester.edu} \\
}

\begin{document}

\usetikzlibrary{shapes.geometric, arrows, positioning}

\tikzset{
  data/.style      = {ellipse, draw, fill=green!30, text width=6em, text centered, minimum height=3em},
  model/.style     = {rectangle, draw, fill=blue!30, text width=7em, text centered, rounded corners, minimum height=3em},
  init/.style      = {rectangle, draw, fill=gray!20, text width=7em, text centered, rounded corners, minimum height=3em},
  trainloop/.style = {rectangle, draw, fill=purple!30, text width=7em, text centered, rounded corners, minimum height=3em},
  measure/.style   = {rectangle, draw, fill=gray!20, text width=7em, text centered, rounded corners, minimum height=3em},
  decision/.style  = {diamond, draw, fill=red!20, text width=5em, text centered, inner sep=0pt, aspect=2},
  sampler/.style   = {rectangle, draw, fill=orange!30, text width=7em, text centered, rounded corners, minimum height=3em},
  arrow/.style     = {thick,->,>=stealth}
}

\maketitle

\begin{abstract}
This study bridges cognitive science and neural network design by examining whether artificial models exhibit human-like forgetting curves. Drawing upon Ebbinghaus' seminal work on memory decay and principles of spaced repetition, we propose a quantitative framework to measure information retention in neural networks. Our approach computes the recall probability by evaluating the similarity between a network's current hidden state and previously stored prototype representations. This retention metric facilitates the scheduling of review sessions, thereby mitigating catastrophic forgetting during deployment and enhancing training efficiency by prompting targeted reviews. Our experiments with Multi-Layer Perceptrons reveal human-like forgetting curves, with knowledge becoming increasingly robust through scheduled reviews. This alignment between neural network forgetting curves and established human memory models identifies neural networks as an architecture that naturally emulates human memory decay and can inform state-of-the-art continual learning algorithms.
\end{abstract}

\section{Introduction}
\label{sec:intro}

Research into the human memory system has been an ongoing endeavor for centuries. One of the most notable early works is Ebbinghaus’ famous experiment, which led to the formulation of the forgetting curve—a mathematical framework describing the probability of recalling learned information over time. His work demonstrated that memory decays exponentially but flattens with the use of spaced repetition review sessions, which significantly improve retention \cite{ebbinghaus}. The pursuit of understanding the intricacies of learning and memory formation has since driven innovations aimed at optimizing learning processes, making them both more efficient and effective. Spaced repetition, for instance, has been repeatedly shown to reduce the decay of memories and enhance their long-term strength \cite{ebbinghaus, Heller1991-eb, ReplicationOfEbbinghaus, Wirth2015-ms, wang2024personalizedforgettingmechanismconceptdriven, Walsh2023-of, Murre2007-uz, Rubin1999-tq}.

Given the extensive knowledge of human memory mechanisms, a natural question arises: \textit{What if we could apply these principles to neural networks—the computational counterparts of the human brain?} 

To explore this idea, the initial step is to identify machine learning models that exhibit human-like forgetting curves—a novel and relatively unexplored research direction. If such models were to be found, state-of-the-art cognitive and neuroscience research could be applied to significantly enhance learning capacity, retention, and efficiency. Moreover, human-like memory models could drive advancements in continual learning algorithms through biologically inspired methods, supporting lifelong learning in non-iid scenarios. For example, replay-based training strategies inspired by REM sleep processes \cite{hayes2021replaydeeplearningcurrent} could be examined on these models to better understand their impact on memory retention. Additionally, dynamic memory buffers could be employed to selectively revisit critical data examples, ensuring effective retention of learned representations.

\clearpage               
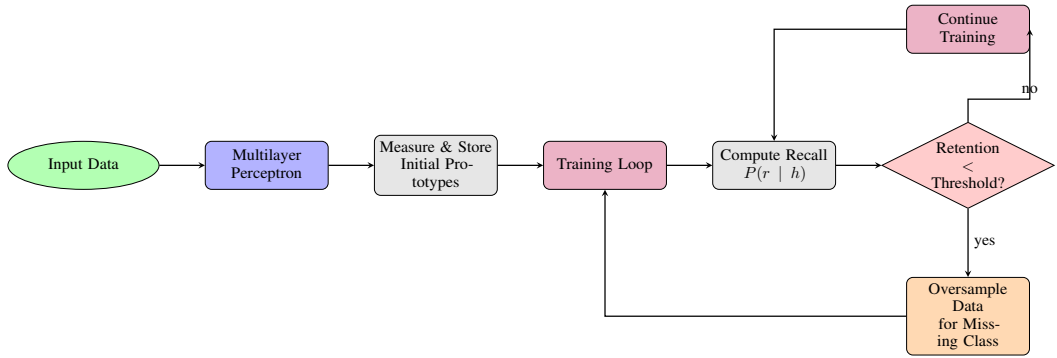
\begin{figure}[!t]       
  \centering
  \resizebox{\textwidth}{!}{%
    \begin{tikzpicture}[node distance=1cm, auto]
    
      \node [data] (data) {Input Data};
      \node [model, right=of data] (mlp) {Multilayer Perceptron};
      \node [init, right=of mlp] (init) {Measure \& Store\\Initial Prototypes};
      \node [trainloop, right=of init] (train) {Training Loop};
      \node [measure, right=of train] (measure) {Compute Recall\\$P(r\mid h)$};
      \node [decision, right=of measure] (check) {Retention\\$<\,$Threshold?};
      \node [sampler, below=1.5cm of check] (oversample) {Oversample Data\\for Missing Class};
      \node [trainloop, above=1.5cm of check] (continue) {Continue Training};
    
      \draw [arrow] (data) -- (mlp);
      \draw [arrow] (mlp) -- (init);
      \draw [arrow] (init) -- (train);
      \draw [arrow] (train) -- (measure);
      \draw [arrow] (measure) -- (check);
      \draw [arrow] (check.south) -- node[right]{yes} (oversample.north);
      \draw [arrow] (oversample.west) -|(train.south) ;
      \draw [arrow] (check.north) -- ++(0,0.5) -| node[above]{no} (continue.east);
      \draw [arrow] (continue.west)  -| (measure.north);
    
    \end{tikzpicture}
  }
  \caption{Retention measuring architecture for spaced-review training loop.}
  \label{fig:spaced-review-arch}
\end{figure}

To rigorously quantify memory retention in neural networks, we propose using the probability of recalling a specific class based on the current hidden state of the network. This hidden state, determined by the layer where the highest-level features are stored—in our MLP experiments, the last hidden layer—is compared against stored class representations. Recall probability is mathematically defined as the similarity between the current hidden state and the learned class representation established during initial learning, normalized by the summation of similarity scores across all stored class representations. These stored representations are sourced either from a sufficiently trained model before further learning or represent the stable plateau of prediction accuracy for a given class. This metric allows for a direct evaluation of memory strength and decay over time.

Finally, to characterize forgetting behavior in neural networks, we will compare experimental outcomes with established mathematical models of human memory. These models include Ebbinghaus' classical exponential and logarithmic forgetting functions \cite{ebbinghaus, ReplicationOfEbbinghaus}, as well as contemporary approaches like the summed exponential function \cite{Heller1991-eb, Rubin1999-tq}.

By leveraging this quantitative framework, we systematically assess neural network forgetting patterns and identify the mathematical models from human memory research that most accurately describe their behavior.

\paragraph{In this work we make the following contributions:}
\begin{enumerate}
\item Identify neural network architectures that exhibit human-like forgetting curves, enabling the application of cognitive and neuroscience insights to enhance current and future learning algorithms.
\item Establish a foundation for continual learning approaches that improve efficiency by reducing retraining requirements while maintaining robust knowledge retention.
\item Introduce a novel quantitative metric for assessing memory strength and decay in neural networks via recall probability of prototype representations.
\item Provide an easy-to-use framework applicable to any neural network system for monitoring training and deployed retention and memory decay.
\end{enumerate}






\section{Background}
\label{sec:background}

This section reviews foundational human memory phenomena—quantitative forgetting curves, spacing and testing effects—and examines their translation into neural network retention strategies, setting the groundwork for our prototype-based recall metric.

\subsection{Human Memory Models}

Early research in experimental psychology established quantitative models of how humans retain and forget information over time. A pioneering contribution was made by Hermann Ebbinghaus, who systematically measured his own memory for lists of syllables and plotted the first forgetting curve in 1885 \cite{Ebbinghaus1885, ReplicationOfEbbinghaus}. Ebbinghaus' forgetting curve demonstrated a characteristic pattern: memory retention drops rapidly soon after learning, then levels off, indicating a progressively slower rate of forgetting for information that survives the initial decline. This basic finding has been replicated and confirmed in subsequent studies \cite{Murre2015}, and it remains one of the most robust empirical generalizations in cognitive psychology. Quantitatively, forgetting curves are often well described by simple mathematical functions – for example, exponential decay or power-law functions – that capture the steep initial loss and gradual long-term retention \cite{Rubin1996}. The precise functional form of the curve has been debated (exponential vs. power-law), but many data sets are well fit by a family of decreasing functions with diminishing loss over time \cite{Rubin1996,Wixted2004}. Regardless of the exact form, the existence of a predictable retention function is a cornerstone of human memory models.

Classical theories of human memory divide memory into multiple systems or stores, each with distinct forgetting dynamics. The influential multi-store model of Atkinson and Shiffrin proposes separate short-term and long-term memory components \cite{Atkinson1968}. Short-term memory (or working memory) has a limited capacity and retains information only briefly unless actively rehearsed, leading to near-complete forgetting of unrehearsed items within seconds (as demonstrated by experiments on short-term recall). Long-term memory, in contrast, can retain information for days, years, or even a lifetime, but is still subject to gradual forgetting in the absence of reinforcement. Forgetting from long-term memory underlies the Ebbinghaus curve described above, whereas forgetting in short-term memory is often much more rapid and is typically attributed to decay or interference processes over extremely short intervals. Subsequent models, such as Baddeley’s working memory framework \cite{Baddeley1974} and Tulving’s distinction between episodic and semantic memory \cite{Tulving1972}, further refine our understanding of memory systems, but they all acknowledge that forgetting is a fundamental property of each system, albeit operating over different time scales.

Notably, forgetting is not viewed simply as a failure of the memory system; contemporary perspectives emphasize its adaptive role. The “new theory of disuse” put forth by Bjork and Bjork posits that memory strength has two facets – a storage strength (how well learned or integrated a memory is) and a retrieval strength (how accessible it is at a given moment) – and that forgetting corresponds to a loss of retrieval strength over time, rather than loss of the stored memory itself \cite{Bjork1992}. According to this theory, a temporary inability to recall something (low retrieval strength) does not imply that all underlying information is erased; in fact, periodic forgetting can be beneficial because subsequent re-learning or retrieval practice can build higher storage strength. 

\subsection{Spacing and Repetition}

One crucial insight from human learning research is that the timing and frequency of review dramatically affect the forgetting curve. In particular, spaced repetition – distributing study or practice sessions over time – has been found to significantly enhance long-term retention compared to massed practice (cramming). This phenomenon, known as the spacing effect, was first noted by Ebbinghaus and subsequently studied in hundreds of experiments over the past century \cite{Cepeda2006}. A comprehensive meta-analysis by Cepeda et al. (2006) reviewed a large body of literature on distributed practice and confirmed that spacing out learning episodes reliably improves recall across a wide range of materials and time scales \cite{Cepeda2006}. When repetitions of the same item are separated by delays (minutes, days, or weeks) rather than presented back-to-back, learners retain the information for much longer intervals. In practical terms, spaced learning flattens the forgetting curve, yielding a slower rate of forgetting after each review, whereas massed learning (repetition without spacing) often leads to quick gains that are rapidly forgotten \cite{Cepeda2008}.

Optimal spacing for learning hinges on the interval between study sessions relative to the target retention interval. Cepeda et al. (2008) \cite{Cepeda2008} showed that, to maximize recall at time T, reviews should be spaced roughly 10–20 percent of T. For instance, an exam in one month benefits from a review a few days later; an exam in one year from a review after one to two months. Gaps that are too short (massed practice) or too long (waiting until nearly total forgetting) both undermine retention, so the inter-study interval must be calibrated to the desired delay.

Beyond fixed spacing, “expanding” schedules—in which the first review occurs soon after learning and each subsequent review is spaced increasingly farther apart—often outperform uniform spacing. Landauer and Bjork (1978) \cite{Landauer1978} found that this approach strengthens memories by catching them just before they would be lost and then challenging them as they stabilize.

The testing effect further magnifies the benefits of spacing: actively retrieving information (e.g., via quizzes or flashcards) enhances consolidation more than passive review. Karpicke and Roediger (2008) \cite{Karpicke2008} demonstrated that repeated, spaced self-testing yields superior long-term retention compared to equivalent restudy time.

In practice, these principles underpin spaced-repetition software (e.g.\ Anki, SuperMemo) and have been validated at scale. Tabibian et al. (2019) \cite{Tabibian2019} analyzed real-world language-learning data, applying machine-learned policies that account for human forgetting dynamics to enhance long-term retention.

Together, these findings—from empirical spacing rules through formal retrieval-practice models to large-scale applications—provide quantitative tools for optimally scheduling learning. They also offer a blueprint for designing neural-network training regimens that mirror human memory decay and reinforcement patterns.

\subsection{Applications to Neural Networks}

Neural networks, unlike biological systems, have no intrinsic weight‐decay mechanism, yet they can abruptly lose previously learned knowledge when trained sequentially on new data—a phenomenon known as catastrophic forgetting \cite{McCloskey1989,Ratcliff1990}. Whereas human memory typically decays gradually along a characteristic forgetting curve, naïve networks update parameters via gradient descent and can overwrite representations for Task A almost immediately after learning Task B. This stark contrast has motivated efforts to endow networks with more human‐like retention dynamics.

One broad family of solutions mimics spaced rehearsal in humans by re-exposing the model to past information during new‐task training. Pseudo-rehearsal generates synthetic exemplars from the network’s own outputs to approximate prior data distributions without storing raw inputs \cite{Robins1995}. More recent replay-based training strategies inspired by REM
sleep processes store a subset of real examples in a memory buffer, and periodically view them \cite{hayes2021replaydeeplearningcurrent}, while generative replay employs a secondary model (e.g., a VAE or GAN) to produce pseudo-samples for replay, successfully preserving performance across many sequential tasks \cite{Shin2017}.

A second approach draws on synaptic consolidation, inspired by hippocampus–cortical consolidation in the brain \cite{McClelland1995}. Elastic Weight Consolidation (EWC) computes a Fisher‐information matrix to identify and penalize changes to parameters critical for old tasks, thus “locking in” essential weights \cite{Kirkpatrick2017}. Synaptic Intelligence accumulates an online importance measure for each weight during training and applies a regularization term to protect those deemed vital \cite{Zenke2017}.

Finally, dynamic‐architecture methods allocate new capacity for novel tasks, preventing interference by freezing existing sub-networks. Progressive Neural Networks instantiate fresh modules for each task, with lateral connections enabling feature transfer while safeguarding previous parameters \cite{Rusu2016}. More recently, spacing‐inspired methods such as “Repeat Before Forgetting” schedule sample revisits based on the model’s own retention estimates—spacing reviews increasingly apart to align with emerging forgetting curves and improve both convergence speed and robustness \cite{Amiri2017}.

These strategies replace abrupt, task‐dependent performance drops with smoother, human‐like retention curves, laying the groundwork for continual‐learning systems that both preserve and adapt knowledge over time.

\section{Method}
In this work, we propose a quantitative measure to assess the probability of recall for a given learned representation based on the hidden state of the neural network. The hidden state is defined as the highest-level feature vector that can be obtained from the network (e.g., the final hidden layer), and the stored prototype are representative vectors for each class. Our approach leverages similarity scores computed between the current hidden state and the stored prototype representation for each category. Specifically, the probability of recall for the correct category is defined as follows:
\begin{equation}
    P(r \mid h) = \frac{\exp\left(\alpha \cdot \text{sim}\left(h, h_{\text{correct}}\right)\right)}{\sum_{c \in \mathcal{C}} \exp\left(\alpha \cdot \text{sim}\left(h, h_c\right)\right)},
    \label{eq:recall_probability}
\end{equation}
where:
\begin{itemize}
    \item $h$ denotes the current hidden state of the network,
    \item $h_{\text{correct}}$ is the prototype representation of the correct category,
    \item $h_c$ represents the prototype for each category $c \in \mathcal{C}$ in the output domain,
    \item $\text{sim}(\cdot, \cdot)$ is a similarity function (e.g., cosine similarity),
    \item $\alpha$ is a scaling parameter that adjusts the sensitivity of the similarity measure.
\end{itemize}

To accurately track each class’s “memory strength,” we extract stored prototypes at two possible points: either from a fully converged network or at the moment when that class’s validation accuracy plateaus. Using these prototypes as anchors, we then build a forgetting curve by evaluating, at every training step, the recall probability defined in Eq.~\eqref{eq:recall_probability}.

Specifically, we train a multi-layer perceptron on our task and periodically compute its recall probability for each class to form a baseline forgetting curve (i.e., without any review). We then introduce spaced-repetition “review sessions” at varying intervals—triggered whenever a class’s recall probability falls below a predetermined threshold—and oversample instances of that class until its recall probability recovers to or exceeds its prior peak. By comparing these experimental curves against classical human-memory models (e.g., the power-law decay of Ebbinghaus \cite{ebbinghaus,ReplicationOfEbbinghaus,Heller1991-eb,Murre2007-uz}), we can quantify how review timing influences retention in neural networks and the similarity to human-like forgetting.

\section{Experiments}
To evaluate the effectiveness of our proposed approach, we design a series of experiments using the MNIST dataset. The experimental design is structured as follows:

\subsection{Datasets and Preprocessing}
The MNIST dataset was selected after considering the need to measure forgetting in the case where classes are not necessarily mutually exclusive in their features. Moreover, the dataset was simple enough for a basic MLP model to fully learn it, thus allowing for better measurement of forgetting curves. Standard preprocessing techniques (e.g., normalization, augmentation) were applied to ensure consistency across experiments.

\subsection{Experimental Setup}
For the baseline MLP model featuring 3 layers of 256 neurons, we perform the following:
\begin{enumerate}
    \item Train the model on the selected dataset until it converges.
    \item On a held out prototype dataset, compute and store the initial recall probabilities for each class using the method defined in Eq.~\eqref{eq:recall_probability}.
    \item On another unseen dataset, begin training the pretrained model using a weighted random sampler, with one or more class weights set to zero to measure retention and memory decay.
    \item Compute the quantitative recall probability defined in Eq.~\eqref{eq:recall_probability} at regular intervals during training.
    \item Generate a baseline forgetting curve by allowing the model to evolve without any review sessions.
    \item Introduce review sessions once memory strength of the held out classes reaches a predetermined level, and record the resulting forgetting curves corresponding to each review schedule.
\end{enumerate}

\subsection{Training Details}

For experimentation, the computational resources include a single RTX 3060-TI Nvidia GPU with 8GB of VRAM, along with an i7-11700KF Intel CPU. The full model tested consisted of 784 features from the flattened MNIST dataset, 10 output classes for the digits, 3 hidden layers with 256 neurons each, batch size of 64 images, and a learning rate of 0.0001.

The PyTorch Lightning architecture was used to create the model, along with two custom Lightning callbacks, one for prototype collection and measurement, and one for calculating when to review a class once a threshold is reached.

\subsection{Results}

For initial testing, we wanted to see the effects of never seeing another example of a class as we train, so we removed all digit \texttt{8} examples from the training set and continued training our fully-trained MLP for 80 additional epochs. 

\begin{figure}[h]
  \centering
  \includegraphics[width=0.7\linewidth]{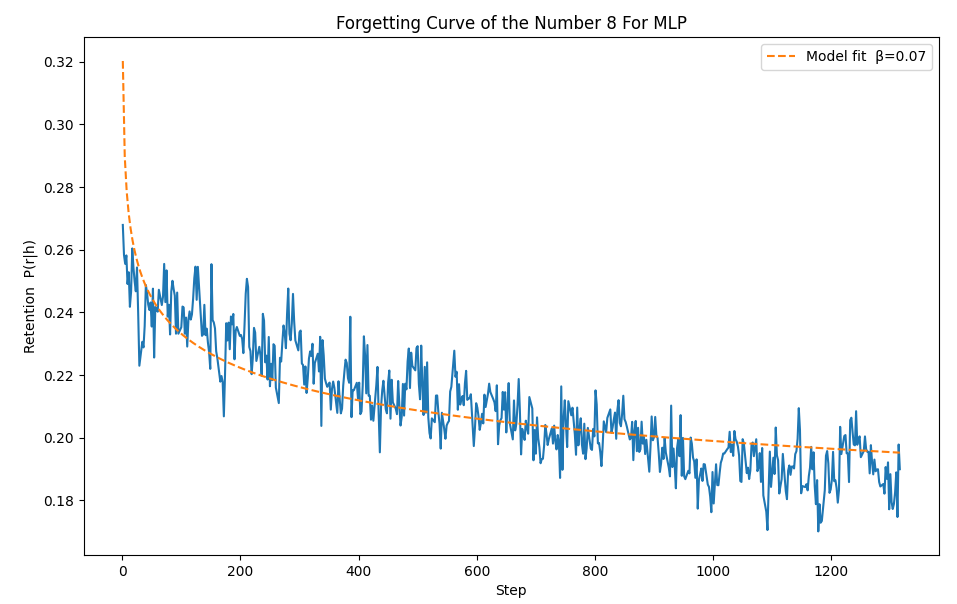}
  \caption{Recall probability of class \texttt{8} over 80 epochs without retraining on \texttt{8}, with a human memory power-law function fit to the retention scores.}
  \label{fig:retention_8}
\end{figure}

As shown in Figure~\ref{fig:retention_8}, the recall probability for class \texttt{8} starts at 0.29 and falls sharply to about 0.20, after which the rate of decline slows. The recall probability began to stabilize around 0.18, rather than 0.0, where the class would be completely forgotten. 

To show the similarity between human memory models and our obtained forgetting curve, we fit a power-law function to the model retention scores. As can be seen in Figure~\ref{fig:retention_8}, the forgetting curve measured from the MLP model closely follows a human-like memory curve.

\begin{figure}[h]
  \centering
  \includegraphics[width=0.7\linewidth]{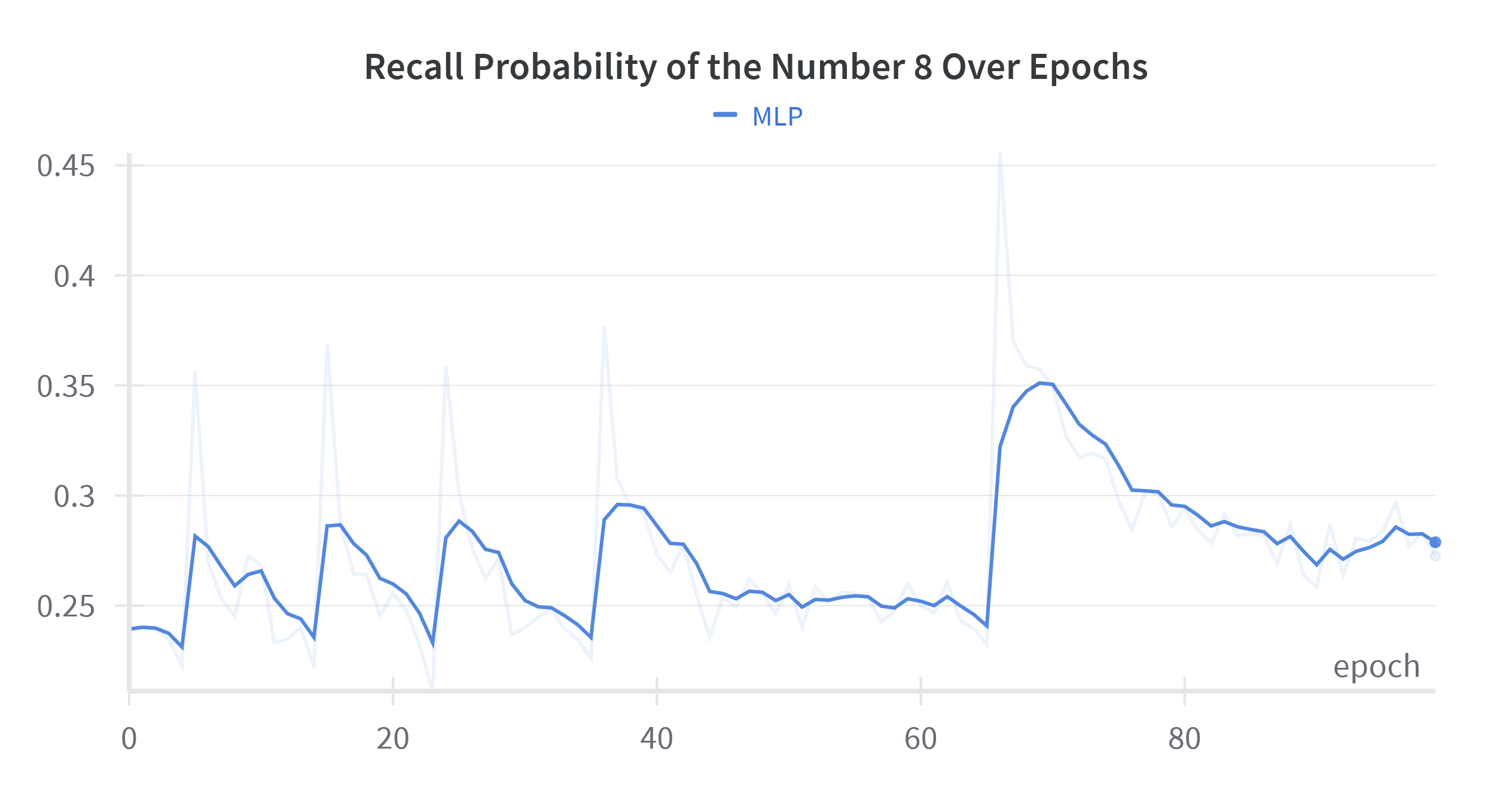}
  \caption{Smoothed recall probability of class \texttt{8} over 100 epochs with 5 spaced reviews.}
  \label{fig:review_8}
\end{figure}

Repeating this protocol, we triggered a “review” of the held-out class as soon as its recall probability dipped to \(\,80\%\) of its original prototype score. Figure~\ref{fig:review_8} shows the smoothed recall probability for class \texttt{8} over 100 epochs, during which five spaced review sessions were needed.

Remarkably, each review immediately restored recall to its pre-review level with no impact on other classes, and the intervals between reviews lengthened progressively—from about 4 epochs before the first review, to 10 before the second and third, and 31 before the fourth. By the fifth review, the peak recall for class 8 exceeded 0.45—more than double the pre-review low—and decayed only gradually through epoch 100.

\section{Discussion}
The relatively low initial recall seen in Figure~\ref{fig:retention_8} can be attributed to the feature overlap between \texttt{8} and other digits (notably the circular strokes in \texttt{6} and \texttt{9}). Similarly, it is hypothesized that the factors keeping the recall probability from reaching close to zero is due to the network's reliance on similar features in other digits to predict the unseen class.

The close alignment of the MLP’s forgetting trajectory with a power-law model suggests that both artificial and biological memory systems may be governed by similar retention dynamics. Importantly, the ability of minimal, well-timed review interventions to not only reverse forgetting but also to extend the intervals of stable recall mirrors the “spacing effect” in human learning, whereby each reinforcement produces more durable memory traces. While these findings offer a compelling analogue to human memory consolidation, our experiments are currently limited to the visual domain and non-mutually exclusive data; further research should assess whether comparable forgetting and recovery patterns emerge in networks trained on time-series data, natural language, or episodic patterns. 

\section{Conclusions}
Our study shows that neural networks such as multilayer perceptrons exhibit human-like forgetting curves and benefit from lightweight, spaced review sessions which can rapidly restore forgotten knowledge in the network and, with repeated reinforcement, progressively fortify the representation against future forgetting. These results not only deepen our understanding of catastrophic forgetting in continual learning but also provide direction on models to focus on, in addition to pointing toward practical strategies—rooted in principles of human memory—for designing review schedules that sustain long-term retention in artificial neural systems.

\section*{Author Contributions}

D.K. contributed to the research as the sole researcher behind the experiments, methods, creation of the manuscript, code, and experimental analysis.

\begin{ack}
We deeply thank fellow researcher Christopher Kanan for his initial insights into this work, and his overall guidance as a professor.
\end{ack}

{
\small
\bibliographystyle{ieee_fullname}
\bibliography{library}

\begin{thebibliography}{10}\itemsep=-1pt

\bibitem{Amiri2017}
Hadi Amiri, Timothy Miller, and Guergana Savova.
\newblock Repeat before forgetting: Spaced repetition for efficient and effective training of neural networks.
\newblock In {\em Proceedings of the 2017 Conference on Empirical Methods in Natural Language Processing (EMNLP)}, pages 2401--2410, 2017.

\bibitem{Atkinson1968}
Richard~C. Atkinson and Richard~M. Shiffrin.
\newblock Human memory: A proposed system and its control processes.
\newblock In {\em The Psychology of Learning and Motivation}, volume~2, pages 89--195. Academic Press, New York, 1968.

\bibitem{Baddeley1974}
Alan~D. Baddeley and Graham~J. Hitch.
\newblock Working memory.
\newblock In Gordon~H. Bower, editor, {\em The Psychology of Learning and Motivation}, volume~8, pages 47--89. Academic Press, New York, 1974.

\bibitem{Bjork1992}
Robert~A. Bjork and Elizabeth~L. Bjork.
\newblock A new theory of disuse and an old theory of stimulus fluctuation.
\newblock In Alan~F. Healy, Stephen~M. Kosslyn, and Richard~M. Shiffrin, editors, {\em Memory: Systems, Process, or Function? (Essays in honor of William K. Estes, Vol. 2)}, pages 35--67. Lawrence Erlbaum Associates, Hillsdale, NJ, 1992.

\bibitem{Cepeda2006}
Nicholas~J. Cepeda, Harold Pashler, Edward Vul, John~T. Wixted, and Doug Rohrer.
\newblock Distributed practice in verbal recall tasks: A review and quantitative synthesis.
\newblock {\em Psychological Bulletin}, 132(3):354--380, 2006.

\bibitem{Cepeda2008}
Nicholas~J. Cepeda, Edward Vul, Doug Rohrer, John~T. Wixted, and Harold Pashler.
\newblock Spacing effects in learning: A temporal ridgeline of optimal retention.
\newblock {\em Psychological Science}, 19(11):1095--1102, 2008.

\bibitem{ebbinghaus}
Hermann Ebbinghaus.
\newblock Memory: A contribution to experimental psychology.
\newblock {\em Ann Neurosci.}, 1885.

\bibitem{Ebbinghaus1885}
Hermann Ebbinghaus.
\newblock {\em Memory: A Contribution to Experimental Psychology}.
\newblock Duncker \& Humblot, Leipzig, 1885.

\bibitem{hayes2021replaydeeplearningcurrent}
Tyler~L. Hayes, Giri~P. Krishnan, Maxim Bazhenov, Hava~T. Siegelmann, Terrence~J. Sejnowski, and Christopher Kanan.
\newblock Replay in deep learning: Current approaches and missing biological elements, 2021.

\bibitem{Heller1991-eb}
O Heller, W Mack, and J Seitz.
\newblock Replikation der ebbinghaus'schen vergessenskurve mit der erspar.
\newblock {\em Replikation der Ebbinghaus'schen Vergessenskurve mit der Erspar}, 1991.

\bibitem{Karpicke2008}
Jeffrey~D. Karpicke and Henry~L. Roediger.
\newblock The critical importance of retrieval for learning.
\newblock {\em Science}, 319(5865):966--968, 2008.

\bibitem{Kirkpatrick2017}
James Kirkpatrick, Razvan Pascanu, Neil Rabinowitz, Joel Veness, Guillaume Desjardins, Andrei~A. Rusu, Kieran Milan, John Quan, Tiago Ramalho, Agnieszka Grabska-Barwi{\'n}ska, Demis Hassabis, Claudia Clopath, Dharshan Kumaran, and Raia Hadsell.
\newblock Overcoming catastrophic forgetting in neural networks.
\newblock {\em Proceedings of the National Academy of Sciences}, 114(13):3521--3526, 2017.

\bibitem{Landauer1978}
Thomas~K. Landauer and Robert~A. Bjork.
\newblock Optimum rehearsal patterns and name learning.
\newblock In M.~M. Gruneberg, P.~E. Morris, and R.~N. Sykes, editors, {\em Practical Aspects of Memory}, pages 625--632, New York, 1978. Academic Press.

\bibitem{McClelland1995}
James~L. McClelland, Bruce~L. McNaughton, and Randall~C. O'Reilly.
\newblock Why there are complementary learning systems in the hippocampus and neocortex: Insights from the successes and failures of connectionist models of learning and memory.
\newblock {\em Psychological Review}, 102(3):419--457, 1995.

\bibitem{McCloskey1989}
Michael McCloskey and Neal~J. Cohen.
\newblock Catastrophic interference in connectionist networks: The sequential learning problem.
\newblock In {\em The Psychology of Learning and Motivation}, volume~24, pages 109--165. Academic Press, 1989.

\bibitem{Murre2007-uz}
Jmj Murre, M Meeter, and A~G Chessa.
\newblock {\em Modeling amnesia: Connectionist and mathematical}.
\newblock 2007.

\bibitem{Murre2015}
Jaap M.~J. Murre and Joeri Dros.
\newblock Replication and analysis of ebbinghaus' forgetting curve.
\newblock {\em PLoS ONE}, 10(7):e0120644, 2015.

\bibitem{ReplicationOfEbbinghaus}
Jaap M.~J. Murre and Joeri Dros.
\newblock Replication and analysis of ebbinghaus’ forgetting curve.
\newblock {\em PLOS ONE}, 10(7):1--23, 07 2015.

\bibitem{Ratcliff1990}
Roger Ratcliff.
\newblock Connectionist models of recognition memory: Constraints imposed by learning and forgetting.
\newblock {\em Psychological Review}, 97(2):285--308, 1990.

\bibitem{Robins1995}
Anthony~V. Robins.
\newblock Catastrophic forgetting, rehearsal and pseudorehearsal.
\newblock {\em Connection Science}, 7(2):123--146, 1995.

\bibitem{Rubin1999-tq}
David~C Rubin, Sean Hinton, and Amy Wenzel.
\newblock The precise time course of retention.
\newblock {\em J. Exp. Psychol. Learn. Mem. Cogn.}, 25(5):1161--1176, Sept. 1999.

\bibitem{Rubin1996}
David~C. Rubin and Amy~E. Wenzel.
\newblock One hundred years of forgetting: A quantitative description of retention.
\newblock {\em Psychological Review}, 103(4):734--760, 1996.

\bibitem{Rusu2016}
Andrei~A. Rusu, Neil~C. Rabinowitz, Guillaume Desjardins, Hubert Soyer, James Kirkpatrick, Koray Kavukcuoglu, Razvan Pascanu, and Raia Hadsell.
\newblock Progressive neural networks.
\newblock {\em arXiv preprint arXiv:1606.04671}, 2016.

\bibitem{Shin2017}
Hanul Shin, Jung~Kwon Lee, Jaehong Kim, and Jiwon Kim.
\newblock Continual learning with deep generative replay.
\newblock In {\em Advances in Neural Information Processing Systems 30 (NeurIPS)}, pages 2990--2999, 2017.

\bibitem{Tabibian2019}
Behzad Tabibian, Utkarsh Upadhyay, Abir De, Ali Zarezade, Bernhard Sch{\"o}lkopf, and Manuel Gomez-Rodriguez.
\newblock Enhancing human learning via spaced repetition optimization.
\newblock {\em Proceedings of the National Academy of Sciences}, 116(10):3988--3993, 2019.

\bibitem{Tulving1972}
Endel Tulving.
\newblock Episodic and semantic memory.
\newblock In Endel Tulving and Wayne Donaldson, editors, {\em Organization of Memory}, pages 381--403. Academic Press, New York, 1972.

\bibitem{Walsh2023-of}
Matthew~M Walsh, Michael~A Krusmark, Tiffany Jastrembski, Devon~A Hansen, Kimberly~A Honn, and Glenn Gunzelmann.
\newblock Enhancing learning and retention through the distribution of practice repetitions across multiple sessions.
\newblock {\em Mem. Cognit.}, 51(2):455--472, Feb. 2023.

\bibitem{wang2024personalizedforgettingmechanismconceptdriven}
Shanshan Wang, Ying Hu, Xun Yang, Zhongzhou Zhang, Keyang Wang, and Xingyi Zhang.
\newblock Personalized forgetting mechanism with concept-driven knowledge tracing, 2024.

\bibitem{Wirth2015-ms}
Michelle~M Wirth.
\newblock Hormones, stress, and cognition: The effects of glucocorticoids and oxytocin on memory.
\newblock {\em Adapt. Human Behav. Physiol.}, 1(2):177--201, June 2015.

\bibitem{Wixted2004}
John~T. Wixted.
\newblock The psychology and neuroscience of forgetting.
\newblock {\em Annual Review of Psychology}, 55:235--269, 2004.

\bibitem{Zenke2017}
Friedeman Zenke, Ben Poole, and Surya Ganguli.
\newblock Continual learning through synaptic intelligence.
\newblock In {\em Proceedings of the 34th International Conference on Machine Learning (ICML)}, volume~70 of {\em Proceedings of Machine Learning Research}, pages 3987--3995, 2017.

\end{thebibliography}
}




\end{document}